\documentclass[letterpaper]{article} 
\usepackage{aaai23}  
\usepackage{times}  
\usepackage{helvet}  
\usepackage{courier}  
\usepackage[hyphens]{url}  
\usepackage{graphicx} 
\usepackage{natbib}  
\usepackage{caption} 
\usepackage{algorithm}
\usepackage{algorithmic}

\usepackage{amsmath}
\usepackage{amssymb}
\usepackage{amsthm}
\theoremstyle{definition}

\usepackage{newfloat}
\usepackage{listings}
\DeclareCaptionStyle{ruled}{labelfont=normalfont,labelsep=colon,strut=off} 
\floatstyle{ruled}
\newfloat{listing}{tb}{lst}{}
\floatname{listing}{Listing}
\nocopyright

\title{Towards a Unified Framework for Sequential Decision Making}
\author{
    Carlos Núñez-Molina, Pablo Mesejo, Juan Fernández-Olivares  
}
\affiliations{
    Universidad de Granada\\

    ccaarlos@ugr.es, pmesejo@ugr.es, faro@decsai.ugr.es
}

\begin{document}

\maketitle


\begin{abstract}

In recent years, the integration of Automated Planning (AP) and Reinforcement Learning (RL) has seen a surge of interest. To perform this integration, a general framework for Sequential Decision Making (SDM) would prove immensely useful, as it would help us understand how AP and RL fit together. 
In this preliminary work, we attempt to provide such a framework, suitable for any method ranging from Classical Planning to Deep RL, by drawing on concepts from Probability Theory and Bayesian inference.
We formulate an SDM task as a set of training and test Markov Decision Processes (MDPs), to account for generalization. We provide a general algorithm for SDM which we hypothesize every SDM method is based on. According to it, every SDM algorithm can be seen as a procedure that iteratively improves its solution estimate by leveraging the task knowledge available. Finally, we derive a set of formulas and algorithms for calculating interesting properties of SDM tasks and methods, which make possible their empirical evaluation and comparison.

\end{abstract}

\section{Introduction}

Throughout the years, two main approaches have been applied to solve Sequential Decision Making (SDM) \cite{littman1996algorithms} problems: Automated Planning (AP) \cite{ghallab2016automated} and Reinforcement Learning (RL) \cite{sutton2018reinforcement}. Even though AP and RL have historically followed separate paths, in recent years there has been a surge of interest in reconciling both approaches. 

Despite this interest, the field currently lacks a comprehensive framework for SDM that suits all existing methods, from Classical Planning to Deep RL. This framework would lay the foundations for a \textit{unified theory of SDM}, providing a deep understanding of SDM and how the different methods fit together, thus helping in the integration of AP and RL. Additionally, it would facilitate the empirical evaluation and comparison of the existing SDM algorithms and tasks.



In this preliminary work, we give the first steps towards this unified framework for SDM, one that suits AP, RL and hybrid methods such as model-based RL \cite{moerland2023model}. In order to provide such framework, we draw on concepts from Probability Theory and Bayesian inference. The main contributions of this framework are the following:


\begin{itemize}
    \item A \textbf{formulation of SDM tasks}, based on the notion of Markov Decision Processes (MDPs) \cite{sutton2018reinforcement}, which works for both AP and RL and takes into account generalization to different MDPs. This is useful for answering questions such as `\textit{Given an AP and an RL task, which elements are similar for both tasks and which ones are different?}' and `\textit{What type of generalization (e.g., across goals) is needed to solve a given task?}'.



    \item A widely applicable definition of the \textbf{difficulty of an SDM task} and distance between MDPs. This makes possible to answer questions such as `\textit{Given an AP and an RL task, which one is harder to solve?}' and `\textit{How different are two particular MDPs?}'.




    \item A \textbf{general algorithm for SDM}, based on Bayesian inference. The main hypothesis of this work is that every SDM method, from both AP and RL, is based on this general algorithm and can be seen as a procedure that iteratively improves its current estimate of the task solution. If our hypothesis is correct, this algorithm 
    would serve as a \textit{template} for creating SDM methods and result useful for answering questions such as `\textit{Given an AP and an RL algorithm, what are their main similarities and differences?}', thus providing valuable insight into SDM.
    



    \item The definitions of several important \textbf{properties of SDM algorithms}: the amount of task knowledge they leverage, how efficiently they solve the task and the quality of the solution found. This allows to answer questions such as `\textit{How much knowledge does a particular algorithm require to solve a task?}' and `\textit{Given an AP and an RL algorithm, which one works better for a particular task?}'.

       
\end{itemize}

To the best of our knowledge, the only other work that provides a unified framework for SDM is \cite{moerland2020framework}. The framework proposed in this work, called \textit{FRAP}, identifies a series of decisions every AP and RL algorithm must make, and provides a comprehensive list of alternatives they can choose from.
On the other hand, our framework formulates SDM from the lens of Probability Theory, based on which it provides a general algorithm for SDM, composed of a sequence of abstract steps that we hypothesize every SDM method follows. Nonetheless, we do not focus on the specific way each algorithm implements (adapts) these steps. Additionally, our framework also provides a novel formulation of SDM tasks and a set of algorithms for calculating interesting properties of SDM tasks and methods, whereas FRAP only focuses on the description of algorithms. Finally, FRAP 
does not cover methods that employ a symbolic knowledge representation (e.g., Classical Planning), while our framework does.

It is also important to note the difference between our framework and Bayesian RL \cite{ghavamzadeh2015bayesian}. This sub-field of RL is composed of methods that explicitly encode their prior knowledge and parameter uncertainty in the form of probability distributions. On the other hand, we propose a general framework in which every SDM algorithm, not only those belonging to Bayesian RL, is formulated as a process which iteratively updates its probabilistic estimate of the solution, either explicitly or implicitly.

\section{Characterization of SDM tasks}

In this section, we propose a general formulation of SDM tasks, suitable for both AP and RL. We draw on concepts from Probability Theory and derive formulas and algorithms to calculate interesting properties and measures, such as the distance between two MDPs and the difficulty of a task.


\subsection{MDP formulation}

We base our formulation on Stochastic Shortest Path Markov Decision Processes (SSP-MDPs) \cite{kolobov2012planning}, a general MDP formulation suitable for both AP and RL, as it encompasses the finite-horizon and infinite-horizon reward-based MDPs usually employed in RL. 
In many tasks, there exist some restrictions which the solution must satisfy (e.g., the maximum amount of fuel each vehicle can consume in a logistics task). Therefore, we augment the SSP-MDP description with a set of constraints, following the ideas introduced by Contrained MDPs \cite{altman1999constrained}. 
We give the name \textit{Constrained Stochastic Shortest Path MDP} (CSSP-MDP) to this hybrid MDP formulation. A CSSP-MDP (which we will simply abbreviate as \textit{MDP} from now on) contains the following elements:

\begin{itemize}
    \item \textbf{State space $S$}, the set of states of the MDP. We define the set of reachable states $S_R$ as the set of states which can be reached by executing a sequence of applicable actions from the initial state of the MDP (see definitions below).
    \item \textbf{Action space $A$}, the set of actions available to the agent. 
    We define the set of applicable actions $App(s) \subset A$ as the subset of actions the agent can execute in some $s \in S$.
    \item \textbf{Transition function $T : S \times A \times S \rightarrow [0,1]$}, a partial function describing the environment dynamics, by specifying the probability $T(s,a,s')$ of the environment transitioning from state $s$ to $s'$ after the agent executes an applicable action $a$ in $s$.
    \item \textbf{Cost function $C: S \times A \times S \rightarrow [0,\infty)$}, a partial function which gives a cost $c(s,a,s') \geq 0$ every time the agent executes an applicable action $a$ in state $s$ and the MDP transitions to state $s'$.
    \item \textbf{Initial state $s_i \in S$}, the state the agent always starts from at the beginning of the task.
    \item \textbf{Goal state $s_g \in S$}, the state the agent must reach. This state is \textit{terminal}, meaning that once the agent is in $s_g$ it cannot get to any other state and no longer incurs in additional cost. Therefore, we can assume the task ends as soon as the agent reaches $s_g$. 
    \item \textbf{Constraint costs function: $D: S \times A \times S \rightarrow \mathbb{R}^K$}, a partial function which returns a vector $(d_1, ..., d_K)$ of \textit{K} real numbers every time the agent executes an applicable action $a$ in state $s$ and the MDP transitions to state $s'$.
    \item \textbf{Constraint values vector $V \in \mathbb{R}^K$}, 
    containing a list of \textit{K} real values, each one associated with a different constraint that must be satisfied by the MDP solution. 
\end{itemize}


A policy $\pi: S \times A \rightarrow [0,1]$ is a partial function which maps 
states $s$ to probabilities over applicable actions $App(s)$. A policy is called \textit{proper} if it always reaches $s_g$, given a sufficiently large number of time-steps. In many SDM formulations, a policy is said to solve an MDP only if it is optimal, i.e., the expected total cost $\mathbb{E}[\![\Sigma c]\!]$ it needs to reach $s_g$ is minimal. Nonetheless, in some SDM tasks (e.g., satisficing planning) the policy is not required to be optimal in order to be considered a solution. Therefore, in our CSSP-MDP formulation, we consider a policy a solution as long as it is proper and it satisfies the $K$ MDP constraints, i.e., for each constraint index $i \in [1, K]$ the expected total constraint cost $\mathbb{E}[\![\Sigma d_i]\!]$ obtained by the policy is less or equal than its corresponding constraint value $v_i$.
Additionally, we assign to each solution policy a quality value $q$ equal to the inverse of its expected total cost: $q=\mathbb{E}[\![\Sigma c]\!]^{-1}$. A policy which does not solve the MDP has a $q$ of 0. These quality values allow us to rank the different solution policies: given two policies $\pi_1, \pi_2$ which solve an MDP $m$ and have qualities $q_1, q_2$, if $q_1 > q_2$ then $\pi_1$ is said to be a better solution of $m$ than $\pi_2$.

After defining when a policy solves an MDP, we now do the opposite and define what the solution of a given MDP is. A first approach could be to consider that the solution of an MDP is simply the set of all policies that solve it. Nonetheless, this definition completely disregards policy quality $q$ so, given two MDPs $m_1, m_2$ which are solved by the same set $\Pi^*$ of policies, the solutions of $m_1$ and $m_2$ would be considered equal even if policies $\pi \in \Pi^*$ obtain different qualities $q^{m_1}, q^{m_2}$ for $m_1$ and $m_2$.
To avoid this, we propose a different definition. Let $\Pi$ be the set of all policies, regardless of whether they solve the MDP or not, which are deterministic, i.e., for every state $s \in S$ they assign a probability of one to some action $a \in App(s)$ and zero to the rest.
We can construct a probability distribution $P^*(\Pi)$ which assigns to each policy $\pi_i \in \Pi$ a probability equal to its quality value $q_i$.\footnote{Throughout the paper, we assume $P^*(\pi_i) = q_i$ but, for this to be true, $q_i$ must first be \textit{normalized} by a constant so that all policy qualities add up to one, i.e., $\sum_{\pi_i \in \Pi} q_i = 1$. Therefore, $P^*(\pi_i)$ is actually equal to the \textit{normalized} quality of $\pi_i$.}
We consider this probability distribution $P^*(\Pi)$ to be the solution of the MDP and thus call it \textit{solution probability distribution}. This formulation has two main advantages. Firstly, it provides a compact description of an MDP solution which takes into account policy quality $q$. Secondly, by formulating the MDP solution as a probability distribution, we can draw on concepts from Probability Theory to characterize SDM tasks and algorithms. Finally, we provide in the Appendix an example where a Classical Planning task is adapted to the scenarios of optimal, satisficing and agile planning by selecting a suitable CSSP-MDP encoding.

\subsection{SDM task formulation}

Most SDM setups define an SDM task in terms of a single MDP. However, we may be interested in solving not one but a set of similar MDPs, e.g., a set of AP problems belonging to the same planning domain (which may have different $s_i, s_g, S$ and $A$\footnote{The state space $S$ and set $A$ of (ground) actions for a given planning problem depend on the problem objects, which may vary across different problems of the same domain.}) or an RL task where a robot is trained in a simulation and then deployed to the real world, where the dynamics $T$ (and other elements) may be different to those of the simulation. If the set of MDPs to solve is very large (or even infinite), it is not feasible to obtain a different policy for each MDP. In this case, we need to obtain a policy which \textit{generalizes} across different MDPs.  Following this idea, we formulate an SDM task as a tuple made up of a set of training MDPs $M_{train}$ and a set of test MDPs $M_{test}$. The goal of an SDM algorithm is to obtain a policy which solves the MDPs in $M_{train}$ and generalizes to similar MDPs, like the ones in $M_{test}$, which are used to evaluate the generalization ability of the policy.

We previously defined a policy $\pi$ as a (partial) function that receives as input the current MDP state $s$ and outputs a probability distribution over actions $a \in App(s)$. Nonetheless, if our goal is to obtain a policy that generalizes across different MDPs, it may be necessary to provide additional information to the policy. For instance, it seems unlikely that a policy $\pi$ is able to generalize to MDPs with different goals $s_g$ unless it knows what the goal $s_g$ of the current MDP to solve is. Therefore, in this case, $\pi$ should receive $s_g$ as input in addition to $s$. 

To account for this extra input information, we propose to use a novel type of policy which we name \textit{Context-Aware (CA) policy}. Given an MDP $m$, a CA-policy (just \textit{policy} from this point onwards) receives as inputs the current state $s$ of $m$ (or an observation of $s$, if the MDP is partially observable), the set of applicable actions $App(s)$ and, optionally, some extra information about $m$ which we refer to as the \textit{MDP context} $\mu$. Then, it outputs a probability distribution over $App(s)$. The MDP context $\mu$ contains information about the MDP elements of $m$, i.e., about $S$, $A$, $T$, $C$, $s_i$, $s_g$, $D$ and/or $V$, and allows the policy to effectively generalize across MDPs which differ for some of these elements. For instance, the policy in our previous example, which received information about $s_g$ in addition to $s$, corresponded to a CA-policy where $\mu=\{s_g\}$. This specific type of CA-policies are known in the literature as goal-conditioned/generalized policies \cite{schaul2015universal}. 
The SDM task must determine the specific MDP context $\mu$ to use, i.e., it must decide what information (about MDP elements) policies will have access to when deciding the next action to execute. For example, in a particular SDM task, the goal may be to find a \textit{standard} policy $\pi(s)$ (i.e., $\mu=\{\}$) that solves $M_{train}$ whereas, in a task where policies must generalize across MDPs with different dynamics $T$, the goal may be to find a CA-policy of the form $\pi(s,T)$, where $\mu=\{T\}$.

There remains the issue of how to encode the information about $s$, $App(s)$ and $\mu$. Different policies may require different encodings of the same MDPs. For example, in RL, a policy may correspond to a convolutional neural network which requires inputs $s, App(s), \mu$ to be provided as images, whereas a policy in AP may require a First-Order-Logic (FOL) representation. To solve this issue, we take inspiration from the division between \textit{problem-space} and \textit{agent-space} proposed in \cite{konidaris2006autonomous}. We allow each task MDP to represent its information as it sees fit. Conversely, each policy must employ an encoding for its inputs ($s, App(s), \mu$) and outputs (action probabilities) that is shared among all MDPs. Then, the SDM task must provide a set of functions to translate policy inputs $s, App(s), \mu$ from the knowledge representation employed by each MDP to the representation of each policy, and another set of functions to translate policy outputs from each policy representation to each MDP representation. We call the former set of functions \textit{perceptual interface} and, the latter, \textit{actional interface}.

\subsection{Difficulty of SDM tasks}

The most important property of an SDM task is arguably its difficulty. We define the difficulty of a task as the sum of its \textit{computational effort}, i.e., how hard it is to find a policy $\pi$ that solves $M_{train}$, and its \textit{generalization effort}, i.e., how hard is for $\pi$ to generalize to $M_{test}$. Additionally, we provide a set of formulas and algorithms to calculate this difficulty, which makes possible to compare a set of tasks to determine which one results more challenging for SDM algorithms.


Firstly, we must define the solution of a set of MDPs. Intuitively, a policy solves a set of MDPs if and only if it solves every single MDP in the set. Therefore, we can consider that the overall quality $q^M$ obtained by a policy $\pi$ on a set of MDPs $M = \{m_1,...,m_n\}$ is equal to the product of the qualities $q^{m_1} ,..., q^{m_n}$ obtained by $\pi$ on each MDP $m_1,...,m_n$ separately. We can then associate a solution probability distribution $P_M^*(\Pi)$ to $M$ given by the following formula: $P_M^*(\Pi) = \prod_{i=1}^n P_{m_i}^*(\Pi)$.\footnote{Probabilities must be normalized so that $\sum_{\pi_i \in \Pi} P_M^*(\pi_i) = 1$.}

After defining the solution $P_M^*(\Pi)$ of a set $M$ of MDPs, we can now define their difficulty $D_M$. Intuitively, $D_M$ measures how hard it is to solve $M$, i.e., to find a solution policy $\pi \in \Pi$ with good quality $q^M$. The easiest case would be when every single policy $\pi \in \Pi$ is optimal. In this case, $P_M^*(\Pi)$ would correspond to a uniform distribution $U(\Pi)$ that assigns the same probability (and quality) to every $\pi \in \Pi$. Building on this fact, we can measure the difficulty $D_M$ of $M$ as how different its solution distribution $P_M^*(\Pi)$ is from the solution distribution $U(\Pi)$ corresponding to the easiest case, i.e., as the \textit{distance} between $P_M^*(\Pi)$ and $U(\Pi)$. By employing the \textit{total variation distance} (TV) \cite{gibbs2002choosing} as our distance metric, we obtain the following formula: $D_M = TV(U(\Pi), P_M^*(\Pi))$.

We can also rely on our probabilistic formulation to measure distances among MDPs. We define the distance $d(M_1, M_2)$ between two sets $M_1, M_2$ of MDPs as the distance between their solution distributions $P_{M_1}^*(\Pi)$, $P_{M_2}^*(\Pi)$: $d(M_1, M_2) = TV(P_{M_1}^*(\Pi), P_{M_2}^*(\Pi))$. Intuitively, this means we consider that $M_1$ and $M_2$ are different if they are solved in a different way, i.e., policies which obtain high quality on $M_1$ exhibit low quality on $M_2$ and vice versa.

Finally, we can calculate the difficulty $D_T$ of an SDM task $T=(M_{train}, M_{test})$ as the sum of its computational effort, which is equal to the difficulty $D_{M_{train}}$ of $M_{train}$, and its generalization effort, which is equal to the distance between $M_{train}$ and $M_{test}$. The formula is as follows:
\begin{multline}
\label{eq:task_difficulty_formula}
D_T = D_{M_{train}} + d(M_{train}, M_{test}) = \\ TV(U(\Pi), P_{M_{train}}^*(\Pi)) +  TV(P_{M_{train}}^*(\Pi), P_{M_{test}}^*(\Pi))
\end{multline}

In order to use Equation \ref{eq:task_difficulty_formula} to obtain the difficulty $D_T$ of a task $T$, we need to be able to compute distances (TV) between probabilities distributions, which may be computationally intractable. Nonetheless, we can approximate these distances. The first distance, $TV(U(\Pi), P_{M_{train}}^*(\Pi))$, measures how different $U(\Pi)$ and $P_{M_{train}}^*(\Pi)$ are. If this distance is small, then $P_{M_{train}}^*(\Pi)$ is similar to $U(\Pi)$, which means that $P_{M_{train}}^*(\Pi)$ distributes qualities evenly across policies, i.e., most policies $\pi \in \Pi$ have near-optimal quality. If this distance is high, then $P_{M_{train}}^*(\Pi)$ is different to $U(\Pi)$ and distributes qualities unevenly across policies, i.e., a few policies have high quality while the rest have zero or near-zero quality. Therefore, as the distance between $P_{M_{train}}^*(\Pi)$ and $U(\Pi)$ becomes larger, the average quality $q^{train}$ obtained by policies on $M_{train}$ decreases, when compared to the quality of the optimal policy. The second distance, $TV(P_{M_{train}}^*(\Pi), P_{M_{test}}^*(\Pi))$, measures how different $P_{M_{train}}^*(\Pi)$ and $P_{M_{test}}^*(\Pi)$ are. If this distance is small, then the solution distributions of $M_{train}$ and $M_{test}$ are similar, meaning that policies $\pi \in \Pi$ tend to obtain similar qualities $q^{train}$, $q^{test}$ on both MDP sets. If this distance is high, then the solution distributions of $M_{train}$ and $M_{test}$ are different, so most policies $\pi \in \Pi$ obtain different $q^{train}$ and $q^{test}$. Based on these ideas, we propose an algorithm based on Monte Carlo sampling \cite{metropolis1949} to estimate $D_T$:

\begin{enumerate}
    \item Uniformly sample a large number $n$ of policies $\pi_1, ..., \pi_n \sim U(\Pi)$.
    This is equivalent to, for each sampled policy $\pi_i$, executing a random action $a \in App(s)$ in each MDP state $s$, while making sure $\pi_i$ always selects the same action $a$ for the same policy inputs $s, App(s), \mu$ (since policies $\pi \in \Pi$ are deterministic).
    \item For each sampled policy $\pi_i$, estimate its quality values $q_i^{train}$ and $q_i^{test}$.
    The quality $q_i^{m_t}$ of $\pi_i$ on some training MDP $m_t \in M_{train}$ can be obtained by executing $\pi_i$ on $m_t$ and then estimating its total expected cost $\mathbb{E}[\![\Sigma c]\!]$ and, for each constraint index $i$, its total expected constraint cost $\mathbb{E}[\![\Sigma d_i]\!]$ as the average of the total costs $\Sigma c$ and total constraint costs $\Sigma d_i$ obtained on $m_t$. $\mathbb{E}[\![\Sigma c]\!]$ and $\mathbb{E}[\![\Sigma d_i]\!]$ are then used to obtain $q_i^{m_t}$. Lastly, $q_i^{train}$ is obtained as the product of the qualities $q_i^{m_t}$ obtained by $\pi_i$ on each $m_t \in M_{train}$. To estimate $q_i^{test}$, we employ the same method but on $M_{test}$. 
    \item $D_T$ can be estimated using the following formula:
    \begin{multline}
    \label{eq:task_difficulty_estimation}
    D_T \approx \Bigl( \frac{1}{n} \cdot \sum_{i=1}^{n} q_i^{train} \Bigr)^{-1} + \Bigl( \frac{1}{n} \cdot \sum_{i=1}^{n} (q_i^{train} - q_i^{test})^2 \Bigr),
    \end{multline}

    where the first term of the sum is an estimate of $D_{M_{train}}$ and the second term an estimate of $d(M_{train}, M_{test})$.
\end{enumerate}

The formula above estimates the difficulty $D_T$ of a task $T$ based on the qualities ($q^{train}$ and $q^{test}$) obtained by policies $\pi \sim U(\Pi)$ on $T$. Therefore, if we want to compare the difficulties $D_{T_1}$, $D_{T_2}$ of two tasks $T_1$, $T_2$, we must make sure qualities have the same scale on both tasks. To ensure this, we can rescale qualities $q^m$ separately for each MDP $m$ in each task $T$, e.g., by dividing every $q^m$ by the maximum obtainable quality in $m$. Additionally, it is important to note that the number $n$ of policies that need to be sampled in order to obtain a good estimate of $D_T$ may be very large. In the Appendix, we propose an approach for improving the estimation of $D_T$ without increasing $n$.

\section{Characterization of SDM algorithms}

In this section, we propose a general formulation of SDM algorithms which suits both AP and RL, in addition to hybrid methods such as model-based RL. We hypothesize all SDM algorithms share some common elements, and use these elements to formulate an abstract, general algorithm for SDM. Finally, we propose a set of properties which can be used to evaluate and compare SDM algorithms, no matter how different, providing formulas and methods for their calculation.

\subsection{SDM algorithm formulation}

The goal of an SDM algorithm is to solve an SDM Task $T=(M_{train}, M_{test})$. In its most abstract form, an SDM algorithm is a procedure that receives some knowledge about the training MDPs $M_{train}$\footnote{The SDM algorithm has no information whatsoever about the test MDPs $M_{test}$, since they are only used to evaluate the generalization ability of the solution (policy) obtained by the algorithm.} of the task and outputs an estimate of the solution $P_{M_{train}}^*(\Pi)$ (simply abbreviated as $P^*(\Pi)$ during this section) of $M_{train}$, corresponding to a (possibly stochastic) policy we hope generalizes to $M_{test}$. 

The type of estimate of $P^*(\Pi)$ depends on the particular algorithm employed. In some cases, the algorithm returns as the estimate of $P^*(\Pi)$ a probability distribution which assigns a probability of 1 to some deterministic policy $\pi \in \Pi$, usually the policy with the best quality $q^{train}$ found by the algorithm, and 0 to the rest. Therefore, the output is a deterministic policy. Some examples of this are \textit{Q-learning} \cite{watkins1992q} and Classical Planning algorithms such as \textit{FastForward} \cite{hoffmann2001ff}. In other cases, the SDM algorithm outputs as an estimate of $P^*(\Pi)$ a probability distribution which assigns non-zero probabilities to more than one policy. Therefore, the output in this case is a stochastic policy, since probability distributions over deterministic policies can be associated with stochastic policies.\footnote{With our formulation, in order to sample actions from a stochastic policy we first need to sample a deterministic policy and, then, use this policy to select the action $a \in App(s)$ to execute in the current state $s$.} Some examples of this can be found in Probabilistic Planning algorithms such as \textit{ProbPRP} \cite{camacho2016fond} and RL algorithms such as \textit{REINFORCE} \cite{williams1992simple}. An interesting example is given by Maximum-Entropy (MaxEnt) RL \cite{haarnoja2018soft}, which tries to optimize both the reward obtained by a policy and its entropy. Hence, MaxEnt-RL algorithms try to estimate $P^*(\Pi)$ as closely as possible, since  $P^*(\Pi)$ represents the stochastic policy which optimizes both reward (cost $C$ in our CSSP-MDP framework) and entropy, i.e., it is the optimal policy under the MaxEnt-RL framework.\footnote{The optimal policy in MaxEnt-RL actually depends on the tradeoff between policy reward and entropy. Nonetheless, we can raise $P^*(\Pi)$ by some constant in order to adapt it to a different tradeoff. For example, the solution distribution 
$P_{M_{train}}^*(\Pi)^2$ gives more weight to reward than entropy, whereas $P_{M_{train}}^*(\Pi)^{0.5}$ gives more weight to entropy.}

Every SDM algorithm requires some knowledge about $M_{train}$ in order to estimate $P^*(\Pi)$. The more knowledge the algorithm employs, the more efficient this estimation will be. This knowledge can be classified according to its \textit{source} (i.e., which MDP element it refers to), \textit{quantity} (i.e., how much knowledge there is available) and \textit{representation} (i.e., how this knowledge is encoded). Given an SDM task, different algorithms can be applied to solve it by adapting the source, quantity and representation of the task knowledge provided.
For example, most AP algorithms require a planning domain (which encodes knowledge about $S$, $A$, $T$, $C$ and possibly other sources) and problem (which encodes knowledge about $S$, $s_i$ and $s_g$), both represented in FOL (dimension \textit{representation}). 
In the case of (model-free) RL, the task knowledge is given in the form of an \textit{environment}, corresponding to either a virtual simulator (e.g., a video game) or the real world (which can be perceived through sensors). At each time step, it provides the current state $s \in S$ of the world, along with $App(s)$. Then, the policy inputs the selected action $a \in App(s)$ into the environment, making it transition from $s$ to some other state $s'$, which is returned alongside the reward $r$. Therefore, the environment provides knowledge about $S$, $A$, $T$, $C$ (and possibly other sources). Intuitively, the quantity of knowledge from each source is smaller than in AP. For this reason, the domain-independent heuristics employed in AP cannot be computed in RL, as the quantity of knowledge provided is not enough. We later show how this value can be estimated.

Given the task knowledge available, how is it leveraged to estimate $P^*(\Pi)$? The answer depends on the specific algorithm employed. For example, AP methods perform a search and reasoning process over the knowledge encoded in the planning domain and problem, whereas RL methods learn from the feedback obtained by interacting with an environment. Nonetheless, despite how diverse SDM algorithms may be, we believe all of them can be formulated as \textit{an iterative process which leverages task knowledge in order to repeatedly improve its current estimate $\hat{P}(\Pi)$ of the solution $P^*(\Pi)$}. Building on this idea, we propose a \textit{general algorithm for SDM}, composed of a sequence of abstract steps which we hypothesize every SDM algorithm follows.


\subsection{A general algorithm for SDM}

This general, abstract algorithm is composed of four main steps: 1) \textit{Initialization to prior distribution}, 2) \textit{Policy sampling and evaluation}, 3) \textit{Policy probability update} and 4) \textit{Update propagation}. We now describe these steps in detail.


\subsubsection{1) Initialization to prior distribution.} As previously stated, an SDM algorithm can be seen as a process that iteratively improves its current estimate $\hat{P}(\Pi)$ of $P^*(\Pi)$. This estimate $\hat{P}(\Pi)$ must be initialized to some probability distribution when the algorithm starts. We call this distribution the \textit{prior solution probability distribution} $P^0(\Pi)$.


Some algorithms start with no prior knowledge about the solution, so $\hat{P}(\Pi)$ is initialized to $U(\Pi)$ (i.e., $P^0(\Pi)=U(\Pi)$). This means that, initially, these algorithms estimate the same solution probability $P^*(\pi)$ (and quality) for every policy $\pi \in \Pi$. A classical example is Q-Learning, which initializes every Q-value $Q(s,a)$ to the same constant value. As a result, initially every action $a \in App(s)$ is equally likely to be sampled for each state $s \in S$, thus making every policy $\pi \in \Pi$ have the same probability $P^0(\pi)=U(\pi)$.

However, other algorithms start with some prior knowledge about the solution, in the form of an initial estimate $P^0(\Pi) \neq U(\Pi)$ of $P^*(\Pi)$. This initial estimate may be used to discard some policies (those $\pi \in \Pi$ for which $P^0(\pi)=0$), which will not be considered by the SDM algorithm, and to decide which policies look more \textit{promising} (those $\pi \in \Pi$ with high $P^0(\pi)$), i.e., more likely to exhibit high quality $q^{train}$ (simply abbreviated as $q$ throughout this section). 
In Deep RL, $P^0(\Pi)$ is mainly determined by the inductive bias \cite{goyal2022inductive} and hyperparameters (e.g., number of layers) used. These two elements limit the expressivity of the neural network employed, meaning that there exist some policies $\pi \in \Pi$ which it \textit{simply cannot represent}. These unlearnable policies $\pi$ are implicitly assigned a prior probability $P^0(\pi)$ of $0$. Therefore, the policy space $\Pi$ to search over is reduced, which often results in more efficient learning of $P_{M_{train}}^*(\Pi)$ and better generalization to $P_{M_{test}}^*(\Pi)$. 
In AP, $P^0(\Pi)$ is often encoded by the heuristic function $h(s)$. Plans (policies) $\pi$
that traverse states $s \in S$ with good (low) heuristic value $h(s)$ are prioritized during the search, meaning that they are implicitly assigned high $P^0(\pi)$. For example, Greedy Best-First-Search expands at each step the node in the \textit{open list} whose associated state $s$ has lowest $h(s)$.

\subsubsection{2) Policy sampling and evaluation.} 
After initializing $\hat{P}(\Pi)$ to $P^0(\Pi)$, the SDM algorithm samples a policy $\pi \sim \hat{P}(\Pi)$, according to a probability given by $\hat{P}(\pi)$, and then evaluates it. Therefore, $\hat{P}(\Pi)$ controls how the search effort is directed. One possibility is to assign high probability $\hat{P}(\pi)$ to a few policies $\pi \in \Pi$ and low probability to the rest. In this case, the SDM algorithm focuses all its search effort on a few promising policies, i.e., those for which it estimates high quality $q$. Alternatively, the probabilities in $\hat{P}(\Pi)$ can be distributed more evenly among all policies $\pi \in \Pi$, so that even less promising policies have a chance of being sampled. The problem of balancing between these two alternative sampling strategies is known in RL by the name of \textit{exploration-exploitation tradeoff}.

Once a policy $\pi \sim \hat{P}(\Pi)$ has been sampled, the SDM algorithm evaluates it in order to obtain its quality $q$. Nonetheless, in many cases only an estimate $\hat{q}$ of the actual policy quality $q$ can be obtained. A typical example is when the MDP is stochastic and the algorithm does not have access to the transition probabilities $T(s,a,s')$. In order to differentiate between a particular estimate $\hat{q}$ of the quality $q$ of some policy $\pi$ and the current belief (estimate) $\hat{P}(\pi)$ of $q$ by the SDM algorithm, we will call $\hat{q}$ a \textit{score} of $\pi$. We assume there exists a \textit{scoring function} $score(\pi)$ which receives as input a policy $\pi$ and outputs a score $\hat{q}$ according to some conditional probability $P(\hat{q}|\pi)$. SDM algorithms implement this scoring function in two main ways:

\begin{itemize}
    \item \textbf{Monte Carlo sampling.} Given a policy $\pi$, the algorithm samples a number $n \geq 1$ of full trajectories (i.e., trajectories from $s_i$ to $s_g$). Then, it estimates the total expected cost $\mathbb{E}[\![\Sigma c]\!]$ and, for each constraint index $i$, the total expected constraint cost $\mathbb{E}[\![\Sigma d_i]\!]$ as the average of the total costs $\Sigma c$ and total constraint costs $\Sigma d_i$ obtained in the trajectories.  $\mathbb{E}[\![\Sigma c]\!]$ and $\mathbb{E}[\![\Sigma d_i]\!]$ are then used to calculate a score $\hat{q}$ of $\pi$. An example method is REINFORCE, which requires full trajectories before estimating the total reward associated with the current policy.

    \item \textbf{Bootstrapping methods.} These methods estimate the quality $q$ of a policy $\pi$ by leveraging other estimates. In RL, this approach receives the name of \textit{Temporal-Difference (TD) Learning}. An example RL method in this category is Q-Learning, where Q-value estimates $Q(s,a)$ are updated based on other estimates $Q(s',a')$.
    The AP algorithm known as \textit{A*} \cite{russell2010artificial} also utilizes bootstrapping, since the evaluation $f(n_p) = g(n_p) + h(n_p)$ of a node $n_p$ can be recursively improved by using the heuristic estimate $h(n_c^*)$ of the best child node $n_c^*$ of $n_p$: $f'(n_p) = g(n_p) + c(n_p ,n_c^*) + h(n_c^*)$, where $c(n_p ,n_c^*)$ is the cost of going from $n_p$ to $n_c^*$. 
\end{itemize}

\subsubsection{3) Policy probability update.} Once the SDM algorithm has evaluated a policy $\pi \sim \hat{P}(\Pi)$ and obtained a score $\hat{q}$, it needs to use this value to update its estimated solution probability $\hat{P}(\pi)$. This probability can be updated by using Bayes Theorem:
\begin{equation}
    \label{eq:bayes_update}
    \hat{P}(\pi|\hat{q}) = \frac{P(\hat{q}|\pi) \cdot \hat{P}(\pi)}{P(\hat{q})}
\end{equation}

The elements in Equation \ref{eq:bayes_update} have the following meaning:

\begin{itemize}
    \item $\hat{P}(\pi)$: current belief (estimate) of the quality $q$ of policy $\pi$ by the algorithm.    
    \item $P(\hat{q})$: how likely is $\hat{q}$ over all policies in $\Pi$.
    \item $\hat{P}(\pi|\hat{q})$: assuming that the algorithm 
    has obtained a score equal to $\hat{q}$ for $\pi$, what should the new estimate $\hat{P}(\pi)$ of the quality $q$ of $\pi$ be?
    \item $P(\hat{q}|\pi)$: assuming that $\pi$ has been sampled from the solution distribution $P^*(\Pi)$, how likely is the algorithm to obtain a score of $\hat{q}$ for $\pi$? We know, by definition, that the solution probability $P^*(\pi)$ of a policy $\pi$ is equal to its quality $q$ (multiplied by a constant). Then, if we assume $\pi$ has been sampled from $P^*(\Pi)$, $\pi$ is likely to have high quality $q$, as policies with high $q$ are more likely to be sampled. We also know that the score $\hat{q}$ corresponds to an estimate of $q$ and, thus, should be directly proportional to $q$ (otherwise, $\hat{q}$ would be misleading). As a consequence of all of this, $P(\hat{q}|\pi)$ is directly proportional to $\hat{q}$.
\end{itemize}

If we assume all score probabilities are very similar, i.e., $P(\hat{q}) \approx p \: \forall \hat{q}$, where $p$ is a constant, we can ignore the $P(\hat{q})$ term in Equation \ref{eq:bayes_update}. Then, we can simplify it to obtain the following new equation:
\begin{equation}
    \label{eq:bayes_update_proportional}
    \hat{P}(\pi|\hat{q}) \propto \hat{q} \cdot \hat{P}(\pi)
\end{equation}
This equation tells us that, given a policy $\pi$ with a score equal to $\hat{q}$, we should increase its estimated solution probability $\hat{P}(\pi)$ if $\hat{q}$ is high and decrease it if $\hat{q}$ is low. This is something every SDM algorithm does. In AP, search effort focuses around promising regions of the state space $S$ so, when a region turns out not to be as promising as initially estimated (i.e., it seems unlikely to lead into a solution), the algorithm directs the search to a different part of $S$. In RL, when the policy $\pi$ executes an action $a$ in a state $s$ which results in high total reward, the algorithm increases the probability of $\pi$ executing $a$ in $s$ again.

\subsubsection{4) Update propagation.}
We have just explained how the score $\hat{q}$ of a policy $\pi$ can be used to update its estimated solution probability $\hat{P}(\pi)$. However, $\Pi$ may contain a very large (or even infinite) number of policies. Therefore, sampling and updating the probability of a single policy $\pi$ at once is very inefficient. In order to overcome this issue, we can harness the information provided by $\hat{q}$ to update not only the probability of a single policy $\pi_i$, but the probabilities of all policies $\pi_1, …, \pi_n \in \Pi$ that are \textit{similar} to $\pi_i$.

We assume there exists a similarity function $similarity(\pi_i,\pi_j)$ used to compute the distance/similarity between two policies $\pi_i, \pi_j$. If the probability $\hat{P}(\pi_i)$ of $\pi_i$ is updated by some quantity $u$, the probability $\hat{P}(\pi_j)$ of another policy $\pi_j$ will be updated by a quantity equal to $u*similarity(\pi_i, \pi_j)$, directly proportional to how similar $\pi_i$ and $\pi_j$ are. For example, if $\hat{P}(\pi_i)$ is increased (or decreased) by $0.1$ and $similarity(\pi_i,\pi_j)=0.5$, then $\hat{P}(\pi_j)$ will be increased (or decreased) by $0.05$.

In order to calculate this similarity function, many SDM algorithms factorize policies $\pi$ into several elements $e_1, …, e_n$, which are then used to measure policy similarity. This means that, in order to assess if two policies $\pi_i=(e_{i1}, …, e_{in})$ and $\pi_j=(e_{j1}, ..., e_{jm})$ are similar, their elements are compared and $similarity(\pi_i, \pi_j)$ is calculated as the number of elements which appear in both policies, normalized to lie in the $[0,1]$ range. For example, assume an SDM algorithm has obtained a score $\hat{q}$ for a policy $\pi_i$ and wants to use $\hat{q}$ to update not only the probability $\hat{P}(\pi_i)$ of $\pi_i$ but of several policies. A deterministic policy $\pi$ (with empty MDP context $\mu$) is a function that maps every state $s \in S$ to some action $a \in App(s)$. Therefore, we can obtain a possible factorization of $\pi_i$ where elements correspond to pairs of the form $(s, \pi_i(s)=a)$, i.e., a state $s \in S$ and the action $a \in App(s)$ selected by $\pi_i$ in $s$. Then, another policy $\pi_j$ will be similar to $\pi_i$ if, for many states $s \in S$, it selects the same action $a$ as $\pi_i$. Therefore, when updating $\hat{P}(\pi_i)$, any other policy $\pi_j \in \Pi$ with at least one \textit{element} in common with $\pi_i$ (i.e., which selects the same action as $\pi_i$ for at least one state) will also see its probability $\hat{P}(\pi_j)$ updated, where the modulus (strength) of the update will be directly proportional to the number of elements in common with $\pi_i$. In RL, this is often done implicitly. Whenever the current policy $\pi_i$ executes an action $a$ in a state $s$ which results in high total reward, the probability of executing $a$ in $s$ again gets increased. When that happens, all policies $\pi_j$ whose factorization contains the element $(s,a)$ (i.e., which select action $a$ in state $s$) also see their probability $\hat{P}(\pi_j)$ increased, including $\pi_i$.

Finally, the four steps explained above give rise to the following \textit{general algorithm for SDM}:


\begin{enumerate}
    \item Initialize the solution estimate $\hat{P}(\Pi)$ to some prior estimate $P^0(\Pi)$ of the task solution $P^*(\Pi)$.
    \item Sample a policy $\pi \sim \hat{P}(\Pi)$ and evaluate it, by using $score(\pi)$ to obtain a score $\hat{q}$ of $\pi$.
    \item Use $\hat{q}$ to update $\hat{P}(\pi)$. A possible way of performing this update is by using Bayes Theorem,
    according to which $\hat{P}(\pi)$ should increase if $\hat{q}$ is high and decrease if $\hat{q}$ is low.
    
    \item Also update the probability of all policies $\pi_1, ..., \pi_n \in \Pi$ which are similar to $\pi$, according to $similarity(\pi_i, \pi_j)$, in an amount proportional to their similarity value.
    \item Repeat steps $2$ to $4$ until $\hat{P}(\Pi)$ becomes a good estimate of $P^*(\Pi)$, and then output $\hat{P}(\Pi)$ as the solution of the SDM task.
\end{enumerate}

The main hypothesis of this work is that \textit{every SDM method is based on this general algorithm and implements steps $1$ to $5$ either explicitly or implicitly}.



\subsection{Properties of SDM algorithms}

We believe SDM algorithms can be characterized according to three main properties: the quantity of knowledge they leverage to solve the task, how efficiently they solve it and the quality of the solution obtained. In this section, we describe these three properties and provide several formulas and algorithms to calculate them. By doing this, we hope to provide a set of tools to evaluate and compare different SDM algorithms (e.g., from AP and RL) in a fair manner, in order to assess which one works better for a particular task.

\subsubsection{Quantity of knowledge.}
Given an SDM algorithm, we may wonder how much knowledge it requires to solve a task. 
Our hypothesis is that all the task knowledge employed by the algorithm is split into three elements: the prior solution distribution $P^0(\Pi)$, the scoring function $score(\pi)$ and the similarity function $similarity(\pi_i, \pi_j)$. If our hypothesis is correct, by measuring the quantity $Q$ of knowledge encoded in these elements, it should be possible to calculate how much task knowledge the algorithm employs.

In order to calculate the quantity of knowledge $Q_{P^0(\Pi)}$ contained in the prior solution distribution $P^0(\Pi)$, we can rely on the formula provided in the previous section for calculating the difficulty $D_M$ of a set $M$ of MDPs: $D_M = TV(U(\Pi), P_M^*(\Pi))$. According to this formula, $D_M$ measures how difficult it is for an SDM algorithm to perfectly estimate $P_M^*(\Pi)$, i.e., reach a situation where it has perfect knowledge about the solution and $\hat{P}(\Pi)=P_M^*(\Pi)$, assuming there exists no prior knowledge about the solution, i.e., $P^0(\Pi)=U(\Pi)$. Nonetheless, in many SDM algorithms there exists a prior solution estimation, i.e., $P^0(\Pi) \neq U(\Pi)$. Consequently, we can adapt the previous formula to calculate the difficulty $D_{M, P^0(\Pi)}$ of $M$ in the presence of some $P^0(\Pi)$: $D_{M, P^0(\Pi)} = TV(P^0(\Pi), P_M^*(\Pi))$. Intuitively, this formula tells us that the prior solution distribution $P^0(\Pi)$ decreases the difficulty of $M$ ($D_{M, P^0(\Pi)} < D_M$), by reducing the existing \textit{knowledge gap} between the initial situation of the SDM algorithm, where $\hat{P}(\Pi)=P^0(\Pi)$, and the final situation of the SDM algorithm, where $\hat{P}(\Pi) \approx P_M^*(\Pi)$, thus reducing the effort required to estimate $P_M^*(\Pi)$.

Given an SDM Task $T=(M_{train}, M_{test})$, we can calculate the quantity of knowledge $Q_{P^0(\Pi)}$ encoded in the prior distribution $P^0(\Pi)$ of some algorithm by measuring to what extent $P^0(\Pi)$ reduces the difficulty $D_{M_{train}, P^0(\Pi)}$ of $M_{train}$. In other words, $Q_{P^0(\Pi)}$ is inversely proportional to $D_{M_{train}, P^0(\Pi)}$ and, thus, we can use it to estimate $Q_{P^0(\Pi)}$. A possible algorithm for estimating $D_{M_{train}, P^0(\Pi)}$, based on Monte Carlo sampling, is the following:

\begin{enumerate}
    \item Sample a large number $n$ of policies $\pi_1, ..., \pi_n \sim P^0(\Pi)$ from the prior distribution $P^0(\Pi)$.
    \item For each sampled policy $\pi_i$, estimate its quality $q_i$ on $M_{train}$.
    \item $D_{M_{train}, P^0(\Pi)}$ can be estimated using the following formula:
    \begin{equation}
    \label{eq:task_difficulty_estimation_with_prior}
    D_{M_{train}, P^0(\Pi)} \approx \Bigl( \frac{1}{n} \cdot \sum_{i=1}^{n} q_i \Bigr)^{-1}
    \end{equation}
\end{enumerate}

Given two algorithms $alg_i$ and $alg_j$ with prior distributions $P_i^0(\Pi)$ and $P_j^0(\Pi)$, respectively, if $D_{M_{train}, P_i^0(\Pi)} > D_{M_{train}, P_j^0(\Pi)}$ then $Q_{P_j^0(\Pi)} > Q_{P_i^0(\Pi)}$.

We now describe a method for calculating the quantity of knowledge $Q_{score}$ encoded in the scoring function $score(\pi)$ of some algorithm.
If $score$ provides good estimates $\hat{q}$ of policy quality $q$, then it should preserve the \textit{quality order} across policies. This means that, given two policies $\pi_i, \pi_j \in \Pi$ with qualities $q_i, q_j$ and scores $\hat{q_i}, \hat{q_j}$, if $q_i > q_j$ then $\hat{q_i} > \hat{q_j}$. Building on this idea, a possible algorithm for estimating $Q_{score}$ is the following:

\begin{enumerate}
    \item Uniformly sample a large number $n$ of policy pairs $(\pi_{11}, \pi_{12}), ..., (\pi_{n1}, \pi_{n2})$, where $\pi_{ij} \sim U(\Pi) \: \forall i,j$.
    \item For each policy pair $(\pi_{i1}, \pi_{i2})$, estimate their qualities $q_{i1}, q_{i2}$ on $M_{train}$ and use $score(\pi)$ to obtain their scores $\hat{q}_{i1}, \hat{q}_{i2}$. For this algorithm to work, the method used to obtain $q_{i1}$ and $q_{i2}$ must provide much better estimates of policy quality than $score(\pi)$.
    
    \item $Q_{score}$ can be estimated using the following formula:
    \begin{equation}
    \label{eq:score_estimation}
    Q_{score} \approx \Bigl( \frac{1}{n} \cdot \sum_{i=1}^{n} \Bigl| \frac{q_{i1}-q_{i2}}{q_{i1}+q_{i2}} - \frac{\hat{q}_{i1}-\hat{q}_{i2}}{\hat{q}_{i1}+\hat{q}_{i2}} \Bigr| \Bigr)^{-1}
    \end{equation}
\end{enumerate}

Intuitively, Equation \ref{eq:score_estimation} compares, for each pair of policies $\pi_{i1}, \pi_{i2}$, their relative difference in quality $q$ (first term of the substraction) with their relative difference in score $\hat{q}$ (second term of the substraction). If both terms are similar for all policies, that means $score(\pi)$ returns good quality estimates $\hat{q}$ and, therefore, $Q_{score}$ is large. Lastly, $Q_{score}$ should be normalized by the time $score(\pi)$ spends to evaluate a policy on average. Otherwise, scoring functions which obtain good quality estimates $\hat{q}$ at the expense of high computational time will have an unfair advantage.

Finally, the quantity of knowledge $Q_{sim}$ encoded in the similarity function $similarity(\pi_i, \pi_j)$ can be estimated in an analogous way to $Q_{score}$. If $similarity(\pi_i, \pi_j)$ provides good estimates of policy similarity, then policies with similar quality $q$ should be assigned high similarity, whereas policies with different $q$ should be assigned low similarity. Therefore, a possible algorithm for estimating $Q_{sim}$ is the following:

\begin{enumerate}
    \item Uniformly sample a large number $n$ of policy pairs $(\pi_{11}, \pi_{12}), ..., (\pi_{n1}, \pi_{n2})$, where $\pi_{ij} \sim U(\Pi) \: \forall i,j$.
    \item For each policy pair $(\pi_{i1}, \pi_{i2})$, estimate their qualities $q_{i1}, q_{i2}$ on $M_{train}$ and similarity value $similarity(\pi_{i1}, \pi_{i2})$.
    
    \item $Q_{sim}$ can be estimated using the following formula:
    \begin{multline}
    \label{eq:similarity_estimation}
    Q_{sim} \approx \Bigl( \frac{1}{n} \cdot \sum_{i=1}^{n} \Bigl| 
    | q_{i1} - q_{i2} | - \\ \bigl(1 - similarity(\pi_{i1}, \pi_{i2} )\bigr)
    \Bigr| \Bigr)^{-1}
    \end{multline}
\end{enumerate}

\subsubsection{Efficiency.}

The efficiency of an SDM algorithm measures the computational resources it needed to solve the task. We propose to measure efficiency in terms of \textit{time} and \textit{space} (i.e., memory), as this is the most widely adopted approach in Computer Science. In RL, nonetheless, it is common to measure the efficiency of algorithms in terms of the amount of data they required to learn the policy, what is known as \textit{data-efficiency}. However, this data-efficiency measure can be decomposed into several elements. Firstly, the quantity of task knowledge $Q$ that is necessary to obtain such data in the first place. For example, in order to obtain samples of the form $(s,a,r,s')$, an environment (e.g., a simulator) with information about $S$, $A$, $T$ and $C$ is required. Secondly, the time and space complexity (efficiency) of the algorithm. For this reason, we believe our proposed method for measuring efficiency suits every SDM algorithm, including RL.

Additionally, we propose to differentiate between the \textit{offline} and \textit{online} efficiency of algorithms. Offline efficiency measures the computational effort needed to estimate $P^*(\Pi)$, i.e., to find a policy $\pi \in \Pi$ that solves the training MDPs $M_{train}$. On the other hand, online efficiency measures the efficiency of the computations required to sample an action $a \in App(s)$ from the policy $\pi$, given the current MDP state $s \in S$. Offline and online efficiency are associated with the Machine Learning concepts of training and inference time, respectively. For example, in AP, offline efficiency would normally measure the time and memory needed to carry out the search for a plan from $s_i$ to $s_g$. In this case, online efficiency would measure the memory needed to store the plan, which is linear in plan length, and time needed to return the next action in the plan, which is $O(1)$. In Deep RL, offline efficiency would mostly correspond to the size of the neural network in memory and the time spent on the backpropagation operations. Similarly, online efficiency would mostly correspond to the size of the neural network in memory and the time needed to perform a forward pass through the network.

\subsubsection{Quality.}

Once the SDM algorithm has finished, it outputs its final estimate of $P_{M_{train}}^*(\Pi)$ in the form of a (deterministic or stochastic) policy $\pi_f = \hat{P}(\Pi)$. Then, $\pi_f$ must be evaluated. As previously commented, we propose to obtain the quality $q_f^{test}$ of $\pi_f$ on the test MDPs $M_{test}$, in order to evaluate its generalization ability. In most situations (e.g., when $M_{test}$ contains stochastic MDPs with unknown dynamics $T$) $q_f^{test}$ will need to be estimated. To perform this estimation, we resort once again to Monte Carlo sampling.

In order to estimate the quality $q_f^{m_t}$ of $\pi_f$ on a single test MDP $m_t \in M_{test}$, we sample several trajectories  with $\pi_f$ on $m_t$. Then, we estimate the total expected cost $\mathbb{E}[\![\Sigma c]\!]$ and, for each constraint index $i$, the total expected constraint cost $\mathbb{E}[\![\Sigma d_i]\!]$ as the average of the total costs $\Sigma c$ and total constraint costs $\Sigma d_i$ obtained in the trajectories. If $\pi_f$ failed to reach the goal state $s_f$ of $m_t$ in some trajectory (given a maximum allowed number of time steps) or some MDP constraint is not satisfied, i.e., $\exists i \: \mathbb{E}[\![\Sigma d_i]\!] > v_i$, then $q_f^{m_t}$ is equal to $0$. Otherwise, $q_f^{m_t} = \mathbb{E}[\![\Sigma c]\!]^{-1}$. Finally, the overall quality $q_f^{test}$ of the policy $\pi_f$ can be obtained as the product of the qualities $q_f^{m_t}$ on all test MDPs $m_t \in M_{test}$.


\section{Conclusions}


In this preliminary work, we have attempted to provide the first fundamentals and intuitions of a unified framework for SDM, suitable for both AP and RL. We have proposed 
a formulation of SDM tasks in terms of training and test MDPs that accounts for generalization, and formulated SDM algorithms as procedures that iteratively improve their current estimate of the solution by leveraging the task knowledge available. In addition, we derived a set of formulas and algorithms for calculating interesting properties of SDM tasks and methods, in order to evaluate and compare them.

In future work, we plan to explore how different tasks and algorithms fit into our framework. In addition, we plan to use the formulas described throughout this work to empirically evaluate and compare these tasks and algorithms, opening the door to hybrid competitions where AP and RL methods compete with each other to solve the same tasks.


\section{Acknowledgements}


We want to thank Dr. Masataro Asai and Prof. Hector Geffner for their valuable feedback during this work.

\bibliography{aaai23}


\appendix

\section{CSSP-MDPs for optimal, satisficing and agile planning}
Let us assume we have a Classical Planning task and want to obtain its CSSP-MDP formulation in three different settings: optimal planning (where plans are required to be optimal), satisficing planning (where plan length is desired but plans are not required to be optimal) and agile planning (where plan length is completely disregarded). We can do this the following way. Firstly, the three CSSP-MDPs share the following elements: $S, A, T, s_i, s_g$. 

To obtain a CSSP-MDP for optimal planning, we assign a cost of 1 (assuming actions have unitary costs) to every transition $(s,a,s')$ where $s \neq s_g$. The constraint costs function $D$ is identical to $C$ and there exists a single constraint whose value $v_1$ is equal to the length of the optimal plan from $s_i$ to $s_g$. This means that, in order for a policy (plan) to be a solution of this MDP, it must be proper and its expected total constraint cost $\mathbb{E}[\![\Sigma d_1]\!]$ (equivalent in this case to the plan length) must be less or equal than $v_1$, i.e., less or equal than the optimal plan length. In other words, only plans of optimal length are solutions of this MDP, as it is the case in optimal planning. The solution distribution $P^*(\Pi)$ of this MDP assigns a probability of 1 to the optimal policy or policies and 0 to the rest.

For satisficing planning, we formulate an CSSP-MDP identical to the one previously used but with no constraints. Now, policies (plans) are required to be proper but not optimal in order to be considered solutions. Additionally, 
since every transition has been assigned a cost of 1, the quality $q$ of each solution policy is equal to the inverse of its plan length. This means that, even though plans are not required to be optimal, shorter plans are considered better than large plans, as it is the case in satisficing planning. The solution distribution $P^*(\Pi)$ of this MDP assigns a probability of 0 to improper policies and a probability inversely proportional to their total cost (plan length) to proper policies.

Lastly, for agile planning, we can reuse the previous CSSP-MDP formulation and only need to change $C$. We now assign a cost of 1 to those transitions $(s,a,s_g)$ which go from any state to the goal state, and assign a cost of 0 to any other transition. With this cost function, all proper policies have the same quality $q=1$. This means that, as long as a plan reaches $s_g$, we do not care about its plan length, as it is the case in agile planning. The solution distribution $P^*(\Pi)$ of this MDP assigns a probability of 0 to improper policies and the same constant probability $p$ to all proper policies.

\section{Efficient estimation of properties of SDM tasks and algorithms}

Throughout this work, we have derived several algorithms (see Equations \ref{eq:task_difficulty_estimation}, \ref{eq:task_difficulty_estimation_with_prior}, \ref{eq:score_estimation}, \ref{eq:similarity_estimation}) 
which sample a number $n$ of policies (either uniformly or from some prior distribution $P^0(\Pi)$) and then leverage their qualities $q$ to estimate different properties of SDM tasks and methods.

In order to perform such an estimation, these algorithms rely on policy quality $q$ being \textit{informative}, i.e., on it being useful to discern between the different policies in $\Pi$. If $q$ is not very informative, this means that most policies $\pi \in \Pi$ have similar or identical quality $q$, with only a few policies exhibiting radically different $q$. This is the case for difficult tasks $T$ (i.e., those with high $D_T$), where most policies $\pi \in \Pi$ either do not solve the task (obtain $q=0$) or obtain a very small and similar quality $q$, and then a small fraction of policies obtain extremely high quality $q$. Conversely, when $q$ is very informative quality values distribute more evenly across all policies in $\Pi$. Then, given two policies $\pi_i, \pi_j \sim U(\Pi)$, the probability that $\pi_i$ and $\pi_j$ have different qualities $q_i$, $q_j$ is high.

The quality of the estimation performed by these algorithms is directly proportional to how informative policy quality $q$ is. Therefore, in cases where $q$ is not very informative, e.g., in difficult tasks, a large number $n$ of policies will need to be sampled in order to obtain a good estimate. For example, let us suppose we want to employ Equation \ref{eq:task_difficulty_estimation} to compare the difficulties of two tasks $T_1$, $T_2$, where $M_{train}$ equals $M_{test}$ in each task. In $T_1$, one in a million policies obtains $q^{train}=1$, while the rest obtain $q^{train}=0$. In $T_2$, one in a billion policies obtains $q^{train}=1$, while the rest obtain $q^{train}=0$. We can see that $T_2$ is much harder to solve than $T_1$ (actually, 1000 times harder). Nonetheless, if we apply our equation to this case by sampling $n=1000$ policies $\pi \sim U(\Pi)$, the probability that none of the sampled policies obtains $q^{train}=1$ is equal to 0.999 for $T_1$ and 0.999999 for $T_2$. In other words, our method will almost surely conclude that $D_{T_1}=D_{T_2}=\infty$, which is wrong. Due to how uninformative $q$ is for both $T_1$ and $T_2$, our method would need to sample and evaluate an extremely large number $n$ of policies (e.g., one million) in order to confidently claim that $T_2$ is more difficulty than $T_1$, which may be computationally intractable in many situations.

We now propose an alternative approach for improving the quality of the estimates returned by our algorithms without increasing the number $n$ of policies that need to be sampled. Let $M$ be the set of MDPs for which we are calculating policy quality $q$. We can transform the solution distribution $P^*(\Pi)$ associated with $M$ in order to obtain another distribution $\tilde{P}^*(\Pi)$ that is \textit{smoother}, i.e., which distributes qualities more evenly across policies $\pi \in \Pi$ than $P^*(\Pi)$. Then, we can calculate policy quality $\tilde{q}$ according to $\tilde{P}^*(\Pi)$ instead of $P^*(\Pi)$. These new qualities $\tilde{q}$ will be more informative than the old qualities $q$, thus reducing the number $n$ of policies that our algorithms need to sample in order to obtain a good estimate.

This transformation from $P^*(\Pi)$ to $\tilde{P}^*(\Pi)$ must preserve the ordering between policies $\pi \in \Pi$ according to their quality. This means that, given two policies $\pi_i, \pi_j \in \Pi$, if $\pi_i$ is better than $\pi_j$ according to $P^*(\Pi)$ (i.e., $q_i > q_j$) then $\pi_i$ must also be better than $\pi_j$ according to $\tilde{P}^*(\Pi)$ (i.e., $\tilde{q}_i > \tilde{q}_j$). This quality transformation that preserves policy ordering is analogous to the RL technique known as \textit{reward shaping} \cite{ng1999policy}. Therefore, in order to obtain $\tilde{P}^*(\Pi)$ from $P^*(\Pi)$, we can employ some reward shaping method to modify the cost function $C$ of the MDPs $m \in M$. Then, we can use this modified cost function $\tilde{C}$ to obtain the \textit{shaped quality} $\tilde{q}$ of any policy $\pi \in \Pi$. For example, let us suppose an MDP $m$ where an agent must go from point $A$ to point $B$. Any policy which reaches $B$ receives $q=1$, whereas policies which do not reach $B$ obtain $q=0$. We can resort to reward shaping to modify the cost function $C$ of $m$ so that, now, policies which do not reach $B$ but get \textit{close} to it obtain a quality $0 < \tilde{q} < 1$ inversely proportional to their final distance from $B$. This new quality $\tilde{q}$ is more informative than the previous quality $q$.

Lastly, it is important to note that, in case we want to compare the estimates obtained on different sets of MDPs (e.g., on $M_{train}$ and $M_{test}$) or SDM tasks, we need to apply the same quality transformation (e.g., the same reward shaping method) to every single MDP set or task. An example is when we utilize Equation \ref{eq:task_difficulty_estimation} to compare the difficulties $D_{T_1}, D_{T_2}$ of two tasks $T_1, T_2$. A quality transformation modifies the solution distributions ($P_{M_{train}}^*(\Pi)$ and $P_{M_{test}}^*(\Pi)$) of a task $T$ and, hence, its difficulty $D_T$. Therefore, it could happen that $T_1$ is more difficulty than $T_2$ but, after applying a different quality transformation to each task, $T_2$ becomes more difficulty than $T_1$. To avoid this, we must apply the same quality transformation to $T_1$ and $T_2$ so that the difficulty ratio $D_{T_1} / D_{T_2}$ remains the same before and after the transformation.

\end{document}